\documentclass{esannV2}
\usepackage{graphicx}
\usepackage[latin1]{inputenc}
\usepackage{amssymb,amsmath,array}
\usepackage{caption}
\usepackage{subcaption}

\begin{document}
%style file for ESANN manuscripts
\title{THDC: Training Hyperdimensional Computing Models with Backpropagation}

%***********************************************************************
% AUTHORS INFORMATION AREA
%***********************************************************************
\author{Hanne Dejonghe and Sam Leroux
%
% DO NOT MODIFY THE FOLLOWING '\vspace' ARGUMENT
\vspace{.3cm}\\
%
% Addresses and institutions (remove "1- " in case of a single institution)
Ghent University - Department of Information Technology\\
Ghent - Belgium
}
%***********************************************************************
% END OF AUTHORS INFORMATION AREA
%***********************************************************************

\maketitle

\begin{abstract}
Hyperdimensional computing (HDC) offers lightweight learning for energy-constrained devices by encoding data into high-dimensional vectors. However, its reliance on ultra-high dimensionality and static, randomly initialized hypervectors limits memory efficiency and learning capacity. Therefore, we propose Trainable Hyperdimensional Computing (THDC), which enables end-to-end HDC via backpropagation. THDC replaces randomly initialized vectors with trainable embeddings and introduces a one-layer binary neural network to optimize class representations. Evaluated on MNIST, Fashion-MNIST and CIFAR-10, THDC achieves equal or better accuracy than state-of-the-art HDC, with dimensionality reduced from 10.000 to 64.
\end{abstract}

\section{Introduction}
\label{section:introduction}

Over the past years, the increased use of edge devices, such as wearables and sensor nodes, has created a need for efficient and robust machine learning solutions that are able to operate with strict memory and energy constraints. Traditional deep learning networks are computationally intensive and memory-heavy, making them ill suited for such resource-limited environments. This challenge has led to an increased interest in alternative paradigms such as hyperdimensional computing (HDC). HDC represents all data as ultra-high-dimensional vectors (hypervectors) \cite{b1}, performing classification using simple vector operations and similarity comparisons \cite{b10}. This makes HDC a perfect fit for edge deployment, offering low-energy inference, inherent robustness to noise, low latency, and lower computational cost. HDC has already been successfully applied to a variety of tasks such as hand gesture recognition \cite{b3}, speech recognition \cite{b4} and robotic control \cite{b6, b7}. For a detailed description of the fundamental concepts, including Item Memory (IM) generation, basic operations, and the baseline training/inference pipeline, the reader is referred to foundational literature on HDC \cite{b1, b10, b3}.

Despite their advantages, HDC models have three key limitations. First, ultra-high dimensionality (typically $D \ge 8000$) is required to ensure\break near-orthogonality between hypervectors and maintain competitive accuracy \cite{b10}. While effective, this increases memory and computational costs, creating a bottleneck for resource-constrained platforms. Prior work has focused on reducing dimensionality without sacrificing performance \cite{b14, b15, b16}.

Second, HDC relies on a randomly initialized, static Item Memory (IM), which cannot adapt or update stored hypervectors, limiting representational capacity. Third, the traditional training process is heuristic: class hypervectors in the Associative Memory (AM) are created by non-adaptively bundling encoded samples \cite{b17}, treating all samples equally regardless of informativeness or redundancy, potentially saturating hypervector dimensions. Various approaches have sought to address these issues \cite{b17, b12, b21, b22, b23}. In short, they either learn only IM or AM, or translate HDC into larger NN architectures. 

In this paper, we introduce Trainable Hyperdimensional Computing (THDC), which tackles all three limitations. THDC replaces the fixed IM with trainable embeddings and employs a single-layer binary neural network (BNN) to learn the AM. By training both IM and AM via backpropagation, THDC enhances model expressiveness and achieves competitive performance with lower-dimensional hypervectors. The source code used for this work, will be made publicly available upon publication.

\section{THDC: Trainable Hyperdimensional Computing}

THDC enables end-to-end training of HDC models via backpropagation, producing compact vector representations. Its training has three objectives: (i) jointly learn hypervectors for the Item Memory (IM), (ii) train class hypervectors for the Associative Memory (AM), and (iii) ensure inference remains identical to the efficient baseline HDC.

\subsection{Architecture}
THDC combines principles of HDC with Binary Neural Networks (BNN). It uses an alternative to static IMs by replacing them with trainable embeddings. This allows the model to adapt to the input data distribution. In a THDC model designed for image classification, two sets of embeddings are optimized jointly: a position memory ($P$) representing the positions of each pixel and a value memory ($V$) representing their values. Each embedding is represented as a trainable matrix. Training $V$ and $P$ via backpropagation allows the model to find a representation subspace where related input is mapped to hypervectors that are closer to each other than random vectors. It leads to an optimal mapping that better preserves information, improving performance at lower dimensionality.

In addition to learning the IM, THDC also learns the AM by training a one-layer BNN, following the approach proposed in \cite{b17}, using their publicly available code \footnote{Code from: https://github.com/sjduan/LeHDC}.

\subsection{Encoding}
THDC uses a simple encoding scheme that flattens an image into a 1D pixel vector. Each pixel is encoded by binding its value hypervector with its positional hypervector, and all pixel encodings are bundled into a single image hypervector. This flattened encoding can be used for grayscale images and RGB images (via channel concatenation). This method has been used in recent work, but in our experience this led to suboptimal performance. However, this is why we propose a different encoding scheme for RGB images.

\begin{figure*}[t!]
    \centering
    \includegraphics[width=1\textwidth]{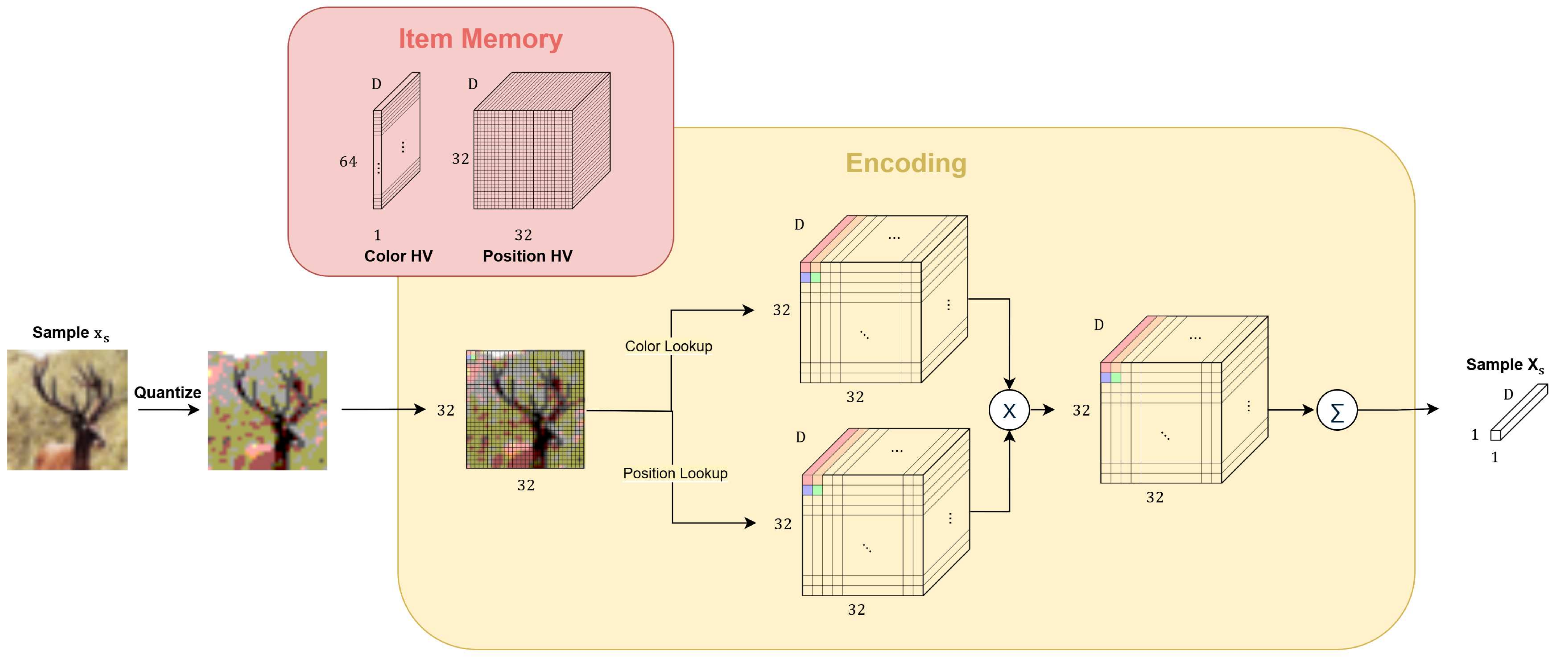}
    \caption{Color encoding scheme used in THDC.}
    \label{fig:encoding}
\end{figure*}

The proposed RGB encoding scheme avoids flattening the channels by first quantizing each pixel's RGB value into one of 64 color bins. Each pixel is then encoded by binding the hypervector of its color-bin with its positional hypervector, followed by bundling all pixel encodings into a single image hypervector, as detailed in Figure \ref{fig:encoding}.

This scheme is expected to perform better for two reasons. First, it preserves semantic color relationships: rather than treating R, G, and B as independent channels, quantization captures the combined color meaning in a single hypervector. Second, reducing the color space to 64 bins, minimizes the size of the IM, which can improve the generalization during training and reduces the model size as additional benefit.

\subsection{Training and Inference}
Training THDC follows the standard pipeline for a one-layer BNN with trainable embeddings. The model starts with two randomly initialized embedding matrices and a binary linear classifier. Each sample is encoded into a hypervector using the current embeddings and the chosen encoding scheme, then passed through the classifier to produce logits.

A cross-entropy loss is computed and gradients are backpropagated through both the classifier and embeddings using the straight-through estimator \cite{b26}.  Classifier weights are binarized in the forward pass, while gradients are taken with respect to their real-valued counterparts, enabling joint updates of the classifier and embeddings. This end-to-end training optimizes the hypervector representations and adapts both IM and AM.

After convergence, the BNN weights are binarized and stored as class hypervectors in the AM, and the IM is taken directly from the learned embeddings, accomplishing objectives (i) and (ii).

During inference, THDC follows the standard HDC pipeline: a query sample is transformed into a hypervector and compared with the class hypervectors in the AM. Thus, although training uses gradient-based optimization, inference stays true to the baseline similarity-based retrieval process and introduces no additional overhead (objective (iii)).

\section{Evaluation}
Our approach aims to improve HDC expressiveness by learning optimal IM and AM hypervectors, reducing the need for ultra-high dimensionality. To demonstrate this and show that we reach competitive accuracy, we evaluate THDC on the MNIST, Fashion-MNIST and CIFAR-10 datasets.

We compare THDC with baseline HDC as described in literature, LeHDC \cite{b17}, a learning-based HDC model using a 2-layer NN \cite{b22} and LDC \cite{b21}. We evaluate these models for $D$ = 32, 64, 100, 1000, 2000, 4000 and 8000. For LDC and the 2-layer NN, we focus on $D$ = 64 based on the results reported in earlier work. Baseline HDC, LeHDC, and THDC use the standard encoding, with THDC additionally tested using the color-bin encoding on CIFAR-10. The others use the encodings defined in their original papers. Training uses cross-entropy loss and the Adam optimizer.

\begin{figure*}[t!]
    \centering
    \includegraphics[width=1\textwidth]{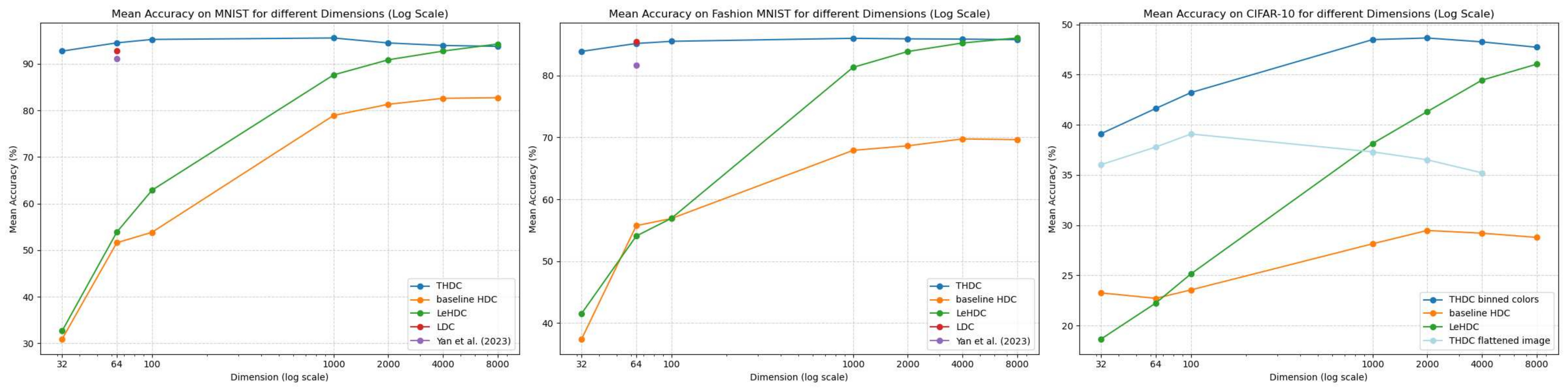}
    \caption{Accuracies for different hypervector dimensions on image datasets.}
    \label{fig:results}
\end{figure*}

Figure \ref{fig:results} reports accuracies across the dimensions on the three datasets. The graph shows that THDC consistently outperforms baseline HDC on all datasets. For instance, at the preferred high dimensionality for baseline HDC of $D$ = 8000 \cite{b24}, THDC achieves an average accuracy increase of 15.38\% compared to baseline. 

THDC matches LeHDC at higher dimensions, but also clearly demonstrates a significant improvement at lower ones. On CIFAR-10, THDC reaches an accuracy of 48.5\% at $D$ = 2000, which is a 2\% increase over LeHDC at its peak performance of $D$ = 8000.  This also is a substantial 20\% improvement over the baseline model at the same dimension. For CIFAR-10, we also compare our two encoding schemes. The color-bin encoding significantly outperforms the flattened image encoding, validating our assumptions and design choice. However, the performance is still not competitive with deep neural networks. It is known that HDC is inferior for complex image recognition tasks.

For the low dimension of $D$ = 64, we compare THDC with the other learning-based baselines that train both IM and AM. As shown in Figure \ref{fig:results}, THDC achieves an accuracy of $94.50$\% which is the best of the three learning-based methods. On Fashion-MNIST, LDC performs only slightly better than THDC. This demonstrates that THDC is effective even at low dimensions and competitive.

\begin{figure}[ht!]
    \centering
    \includegraphics[scale=0.13]{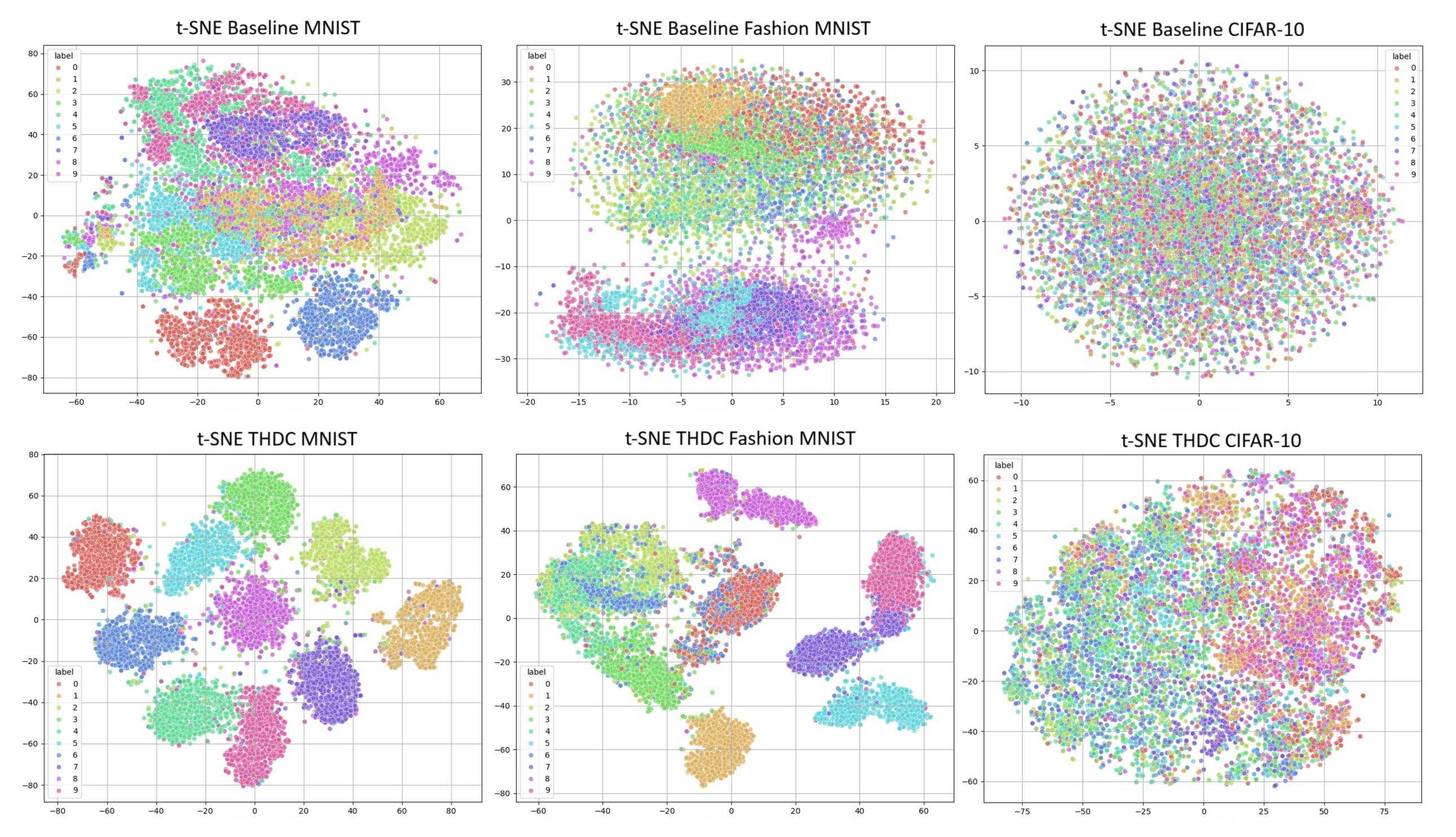}
    \caption{t-SNE plots comparing baseline HDC (up) and THDC (down) for $D=1000$ on MNIST (left), Fashion-MNIST (middle), and CIFAR-10 (right).}
        \label{fig:tsne_comparison}
\end{figure}

To assess the benefit of learning IM vectors, we examine hypervector quality at $D=1000$ via a t-distributed Stochastic Neighbor Embedding (t-SNE) plot. Comparing representations created using a random IM (baseline HDC) with those from our trained IM (THDC), the learned embeddings produce clearer class clusters on MNIST and Fashion-MNIST, whereas the baseline shows substantial overlap. On CIFAR-10, both methods yield less separation, but THDC still shows visibly improved clustering (Figure \ref{fig:tsne_comparison}).

\section{Conclusion}
In this work, we introduced Trainable Hyperdimensional Computing to address key limitations of traditional HDC. By jointly learning the IM through trainable embeddings and the AM with a single-layer BNN, we can overcome these limitations. We also demonstrated the importance of encoding, using a color binning approach for CIFAR-10. Our results show that THDC consistently outperforms baseline HDC and t-SNE plots confirm that learning the IM, aids in mapping data to optimal representations. Furthermore, THDC achieves an accuracy of 48.48\% on CIFAR-10 with only 1000 dimensions and is on par with other learning-based HDC models for $D$ = 64, succeeding in eliminating the ultra-high dimensionality of HDC and making it more suitable for inference on edge-devices. In future work, it is worth further investigating the combination of learning-based methods and HDC as it shows significant improvements to baseline HDC. This might be a way to bridge the gap with the more performant neural networks. Moreover, the search for a data-specific encoding scheme remains important even when learning the IM and could benefit from further research.

% ****************************************************************************
% BIBLIOGRAPHY AREA
% ****************************************************************************

\begin{footnotesize}

\end{footnotesize}


\begin{thebibliography}{99}

\bibitem{b1} P. Kanerva, ``Hyperdimensional computing: An introduction to computing in distributed representation with high--dimensional random vectors,'' \emph{Cognitive Computation}, vol. 1, no. 2, pp. 139--159, 2009.

\bibitem{b10} D. Kleyko, D. A. Rachkovskij, E. Osipov, and A. Rahimi, ``A survey on hyperdimensional computing aka vector symbolic architectures, part I: Models and data transformations,'' \emph{ACM Computing Surveys}, vol. 55, no. 6, pp. 1--40, Dec. 2022.


\bibitem{b3} A. Rahimi et al., ``Hyperdimensional biosignal processing: A case study for EMG-based hand gesture recognition,'' in \emph{2016 IEEE ICRC}, 2016, pp. 1--8.


\bibitem{b4} M. Imani, D. Kong, A. Rahimi, and T. Rosing, ``VoiceHD: Hyperdimensional computing for efficient speech recognition,'' in \emph{2017 IEEE ICRC}, 2017, pp. 1--8.

\bibitem{b6} A. Mitrokhin, P. Sutor, C. Fermüller, and Y. Aloimonos, ``Learning sensorimotor control with neuromorphic sensors: Toward hyperdimensional active perception,'' \emph{Science Robotics}, vol. 4, 2019.

\bibitem{b7} P. Neubert, S. Schubert, and P. Protzel, ``An introduction to hyperdimensional computing for robotics,'' \emph{KI--Künstliche Intelligenz}, vol. 33, no. 4, pp. 319--330, 2019.

\bibitem{b14} T. Basaklar, Y. Tuncel, S. Y. Narayana, S. Gumussoy, and U. Y. Ogras, ``Hypervector design for efficient hyperdimensional computing on edge devices,'' 2021.

\bibitem{b15} J. Morris et al., ``CompHD: Efficient hyperdimensional computing using model compression,'' in \emph{2019 IEEE/ACM ISLPED}, 2019, pp. 1--6.

\bibitem{b16} M. Imani et al., ``SparseHD: Algorithm-hardware co-optimization for efficient high-dimensional computing,'' in \emph{2019 IEEE 27th Annual International Symposium on Field-Programmable Custom Computing Machines}, 2019, pp. 190--198.

\bibitem{b17} S. Duan, Y. Liu, S. Ren, and X. Xu, ``LeHDC: Learning-based hyperdimensional computing classifier,'' in \emph{Proceedings of the 59th ACM/IEEE DAC}, 2022, pp. 1111--1116.

\bibitem{b12} J. Kim, H. Lee, M. Imani, and Y. Kim, ``Efficient hyperdimensional learning with trainable, quantizable, and holistic data representation,'' in \emph{2023 DATE}, 2023, pp. 1--6.

\bibitem{b21} S. Duan, X. Xu, and S. Ren, ``A brain-inspired low-dimensional computing classifier for inference on tiny devices,'', 2022.

\bibitem{b22} Z. Yan et al., ``Efficient hyperdimensional computing,'' in \emph{Machine Learning and Knowledge Discovery in Databases: Research Track}, 2023, pp. 141--155.

\bibitem{b23} D. Ma, C. Hao, and X. Jiao, ``Hyperdimensional computing vs. neural networks: Comparing architecture and learning process,'' in \emph{2024 25th International Symposium on Quality Electronic Design}, 2024, pp. 1--5.

\bibitem{b26} M. Courbariaux, Y. Bengio ``BinaryNet: Training Deep Neural Networks with Weights and Activations Constrained to +1 or -1,'' in \emph{CoRR}, 2016.

\bibitem{b2} H. Amrouch et al., ``Brain-inspired hyperdimensional computing for ultra-efficient edge AI,'' in \emph{2022 CODES+ ISSS}, 2022, pp. 25--34.

\bibitem{b24} M. Imani \emph{et al.}, ``QuantHD: A quantization framework for hyperdimensional computing,'' \emph{IEEE Transactions on Computer-Aided Design of Integrated Circuits and Systems}, vol. 39, no. 10, pp. 2268--2278, 2020.

\end{thebibliography}
\end{document}